\documentclass[letterpaper, 10 pt, conference]{ieeeconf}  

\IEEEoverridecommandlockouts                              

\usepackage{times}
\usepackage{epsfig}
\usepackage{graphicx}
\usepackage{grffile}
\usepackage{amsmath}
\usepackage{amssymb}
\usepackage{multirow}
\usepackage{subfigure}
\usepackage{hyperref}

\title{\LARGE \bf
Dynamic Event Camera Calibration
}

\author{Kun Huang$^{1}$, Yifu Wang$^{1}$ and Laurent Kneip$^{1}$
\thanks{$^{1}$ShanghaiTech University; L. Kneip is also with the Shanghai Engineering Research Center of Intelligent Vision and Imaging. The authors would like to acknowledge the support given by the Natural Science Foundation of Shanghai (grant number: 19ZR1434000).}%
\thanks{Video: \url{youtu.be/ipEM9QxKYO8}}
\thanks{Code: \url{github.com/MobilePerceptionLab/EventCalib}}
}

\begin{document}
\bstctlcite{IEEEexample:BSTcontrol}

\maketitle
\thispagestyle{empty}
\pagestyle{empty}

\begin{abstract}

Camera calibration is an important prerequisite towards the solution of 3D computer vision problems. Traditional methods rely on static images of a calibration pattern. This raises interesting challenges towards the practical usage of event cameras, which notably require image change to produce sufficient measurements. The current standard for event camera calibration therefore consists of using flashing patterns. They have the advantage of simultaneously triggering events in all reprojected pattern feature locations, but it is difficult to construct or use such patterns in the field. We present the first dynamic event camera calibration algorithm. It calibrates directly from events captured during relative motion between camera and calibration pattern. The method is propelled by a novel feature extraction mechanism for calibration patterns, and leverages existing calibration tools before optimizing all parameters through a multi-segment continuous-time formulation. As demonstrated through our results on real data, the obtained calibration method is highly convenient and reliably calibrates from data sequences spanning less than 10 seconds.

\end{abstract}

\section{Introduction}

Over the past decade, we have seen the emergence of several intelligent mobile devices such as smart vehicles, intelligence augmentation devices, or factory and service automation robots. Such devices need to move either actively or passively in the real world and gain an understanding of the geometry of the environment all while keeping track of location. The resulting problem is primarily a geometric perception problem the solution to which is commonly achieved using simple cameras. Although a large body of research has already lead to a certain level of maturity, vision-based methods keep facing challenges in scenarios with high dynamics, low texture distinctiveness, or challenging illumination conditions. Dynamic vision sensors---also called event cameras---present an interesting and innovative alternative in this regard. They independently measure pixel-level logarithmic brightness changes at high temporal resolution and high dynamic range. The advantages and challenges of event-based vision are well explained by the original work of Brandli~et~al.~\cite{brandli2014240} as well as the recent survey by Gallego~et~al.~\cite{gallego2019event}. The present work addresses intrinsic calibration of event cameras, a fundamental problem affecting potential future use in 3D vision applications such as ~\cite{rebecq2018emvs,rosinol18,zhu2018realtime,zhou2021event,jiao2021CVPR,xin2020GOECM,xin2021GOCME,wang2021visual}.

\begin{figure}[t]
  \centering
  \includegraphics[width=0.8\linewidth]{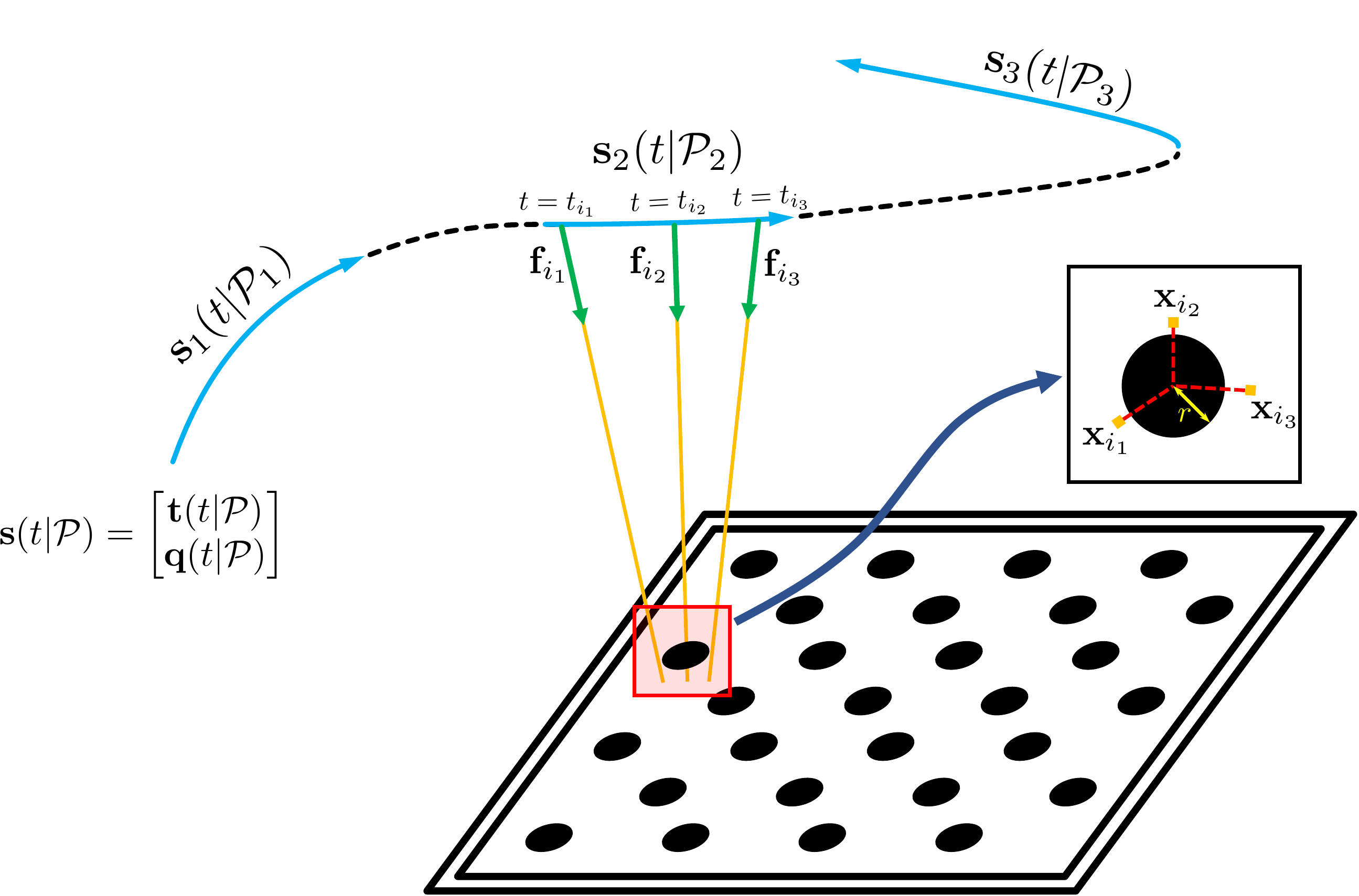}
  \caption{Concept of our event camera calibration framework. The pattern is rendered visible by keeping the camera under motion. Well-distributed sub-segments of the trajectory are chosen and optimized along with the intrinsic camera parameters.}
  \label{fig:concept}
\end{figure}

The physical structure of an event camera is entirely comparable to a regular camera as it is composed of a lens in front of an imaging plane. The efficient usage of event cameras towards the solution of 3D reconstruction problems therefore requires the prior identification of a similar set of intrinsic calibration parameters (e.g. focal length, principal point, etc.). For regular cameras, such calibration procedures are commonly realized by using static images captured in front of a planar calibration target with known features. These features are then re-identified in each image, from which we can obtain a sufficient number of 2D-to-2D correspondences to identify the pose of each view as well as the intrinsic parameters. The problem with calibrating an event camera is that---while such calibration procedures are entirely mature---they are not immediately applicable to event streams for the obvious reason that event cameras do not produce any information when neither camera nor pattern moves. In order to produce frame-like images from events that readily permit the quasi-instantaneous reidentification of pattern feature reprojections under an identical camera pose, the calibration of event cameras requires a modified pattern with simultaneously flashing features. Even if kept still, local accumulations of events can then be used to detect the points of interest and create the necessary input data for reusing traditional calibration methods.

Although this idea sounds relatively straightforward, it suffers from two important practical drawbacks:
\begin{itemize}
  \item Manipulation: The most straightforward way of obtaining a flashing pattern is to visualize the pattern on a screen. However, it is sometimes desirable to move the pattern rather than the camera, which makes the usage of screens impracticable. It is furthermore difficult to ensure mechanical properties of a screen such as flatness and stiffness, especially as the screen becomes larger.
  \item Construction: An alternative is given by using an array of LEDs that are triggered simultaneously by the same circuitry. While such constructions are possible, they are hard and expensive to produce given the high accuracy requirements on the placement of the LEDs.
\end{itemize}

We present \textit{dynamic event camera calibration}, a convenient framework which relies on the following insights and techniques:
\begin{itemize}
  \item We show that it is in fact unnecessary to capture a flashing target from static camera views. A proper choice of a regular calibration target lets us easily design a robust feature extractor that works as soon as the camera or pattern is under motion.
  \item After initialization using off-the-shelf techniques, our calibration concludes with a continuous-time trajectory-fragment based motion compensation scheme in which each event is accounted for with its individual time stamp.
  \item The camera model can be flexibly interchanged, thus permitting the calibration of perspective cameras with or without distortion as well as more exotic fisheye, omni-directional, or catadioptric cameras.
\end{itemize}
Our method has the following advantages:
\begin{itemize}
  \item It relies on a regular calibration pattern. No extra efforts for making a calibration target are required.
  \item The camera or calibration target can be moved at considerable speed, thus permitting accurate and convenient calibration from data sequences as short as 10s.
  \item Owing to the nature of event cameras, little attention has to be paid to illumination conditions.
  \item Owing to the fact that we use a common calibration target, the method easily permits the extrinsic calibration to other, possibly regular cameras.
\end{itemize}
Our fully integrated, ready-to-use event camera calibration framework builds upon existing tools and will be released to the community.

Our paper is structured as follows. Section \ref{sec:relatedwork} reviews further related work. Section \ref{sec:main} then gives an overview of our calibration framework including the details of our feature extraction technique as well as the continuous-time back-end optimization technique. Section \ref{sec:experiments} finally concludes with numerous tests on real camera-lens combinations.

\section{Related work}
\label{sec:relatedwork}

Camera calibration is an important topic in geometric vision. The various presented methods depend strongly on the employed camera projection model, the most prominent of which is the perspective camera model \cite{hartley04}. Though alternative methods exist \cite{sturm04}, the most prominent camera calibration method for perspective cameras was introduced by Zhengyou Zhang \cite{zhang99,zhang20}, a technique which found its way into Bouguet's highly popular calibration toolbox \cite{bouguet} as well as a related OpenCV \cite{opencv_library} implementation. These frameworks are able to handle perspective cameras both with and without distortions, and have furthermore served as a foundation for other toolboxes that are able to calibrate fisheye and catadioptric cameras~\cite{scaramuzza06}.

As mentioned in the introduction, the current standard event camera calibration method \cite{ev1,ev2,ev3} consists of using a flashing calibration pattern. Simple short term accumulations of events complemented by a corner extraction method can deliver the instantaneous corner locations of the pattern, and thus enable the reuse of all above-mentioned calibration toolboxes. In contrast, our proposed method is able to use conventional camera calibration targets.

Our procedure relies on feature detection in event streams. Several techniques for event-based corner extraction have already been presented~ \cite{clady15,vasco16,mueggler17,alzugaray18,alzuguray18a2}, and they could be used for detecting the corners of a checkerboard pattern. However, as we will explain, it is advantageous to use a grid pattern of circles. Our feature extraction method hence draws analogies with the event-based ball detection presented by Glover et al. \cite{glover16}. It relies on noise-resilient cluster detection \cite{ester1996density} as well as circle fitting \cite{kaasa1976circle}.

The core of our contribution is given by a back-end motion compensation framework that relies on a continuous-time parametrization of the camera motion. The employed representation shares analogies with the B-spline-based~\cite{piegl2012nurbs} framework presented by Furgale et al. \cite{furgale2015continuous}, and is complemented by the cubic spline interpolation technique presented in \cite{kang1999cubic} as well as the B-spline adaptations to Lie groups presented by Sommer et al. \cite{sommer2016continuous} and Sommer et al. \cite{sommer2020efficient}. Another strongly related work is given by the rolling shutter camera calibration method presented by Oth et al.~\cite{oth13}, who also use a continuous-time parametrization to calibrate camera-inherent parameters. Continuous-time parametrizations have also been used for event-based camera localization~\cite{mueggler15,rosinol18}. However, we are---to the best of our knowledge---the first to apply the model to intrinsic event camera calibration.


\section{Dynamic event camera calibration}
\label{sec:main}

We present a novel procedure for calibrating event cameras using only a regular calibration pattern. More specifically, our method employs patterns that consist of a square grid of regularly spaced black, circular dots. The intuition behind this choice is that such features---owing to their radially symmetric nature---may be robustly extracted using a rotation-invariant feature detector as soon as local optical flow exceeds a certain level. More specifically, the event patterns generated by the circular dots are similar up to a rotation in the image plane, and notably given by two point-symmetric clusters of positive and negative events appearing at opposing poles of each circle. We exploit this fact in order to design an efficient circle extraction algorithm. The section is organized as follows. Section \ref{sec:framework_overview} provides an overview of our complete calibration framework. Section \ref{sec:feature extraction} presents our first important sub-module given by the feature extraction. Section \ref{sec:optimization} finally concludes with our novel continuous-time back-end optimization module that refines the initial result over both identified intrinsic parameters as well as the dynamic motion parameters.


\begin{figure}[ht!]
  \centering
  \vspace{0.2cm}
  \includegraphics[width=0.9\linewidth]{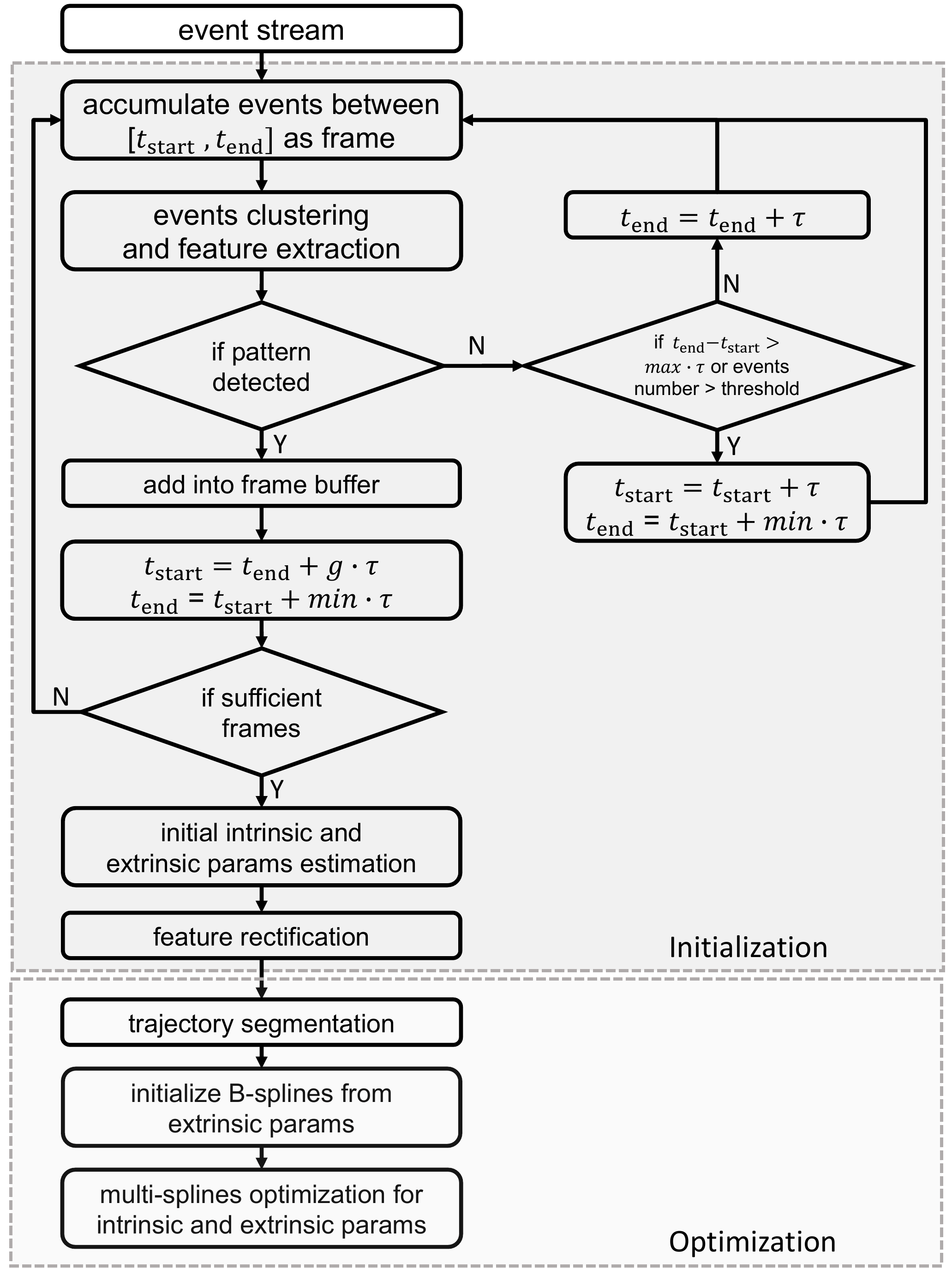}
  \caption{Block diagram of our method containing both the front-end initialization and the back-end optimization part.}
  \vspace{-0.2cm}
  \label{fig:flowchart}
\end{figure}

\subsection{Framework overview}
\label{sec:framework_overview}

\begin{figure}[b]
\centering
\includegraphics[width=0.45\linewidth]{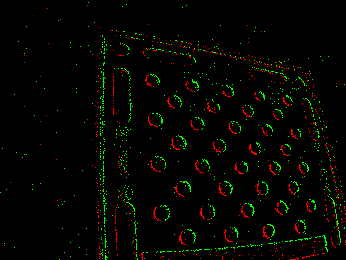}
\includegraphics[width=0.45\linewidth]{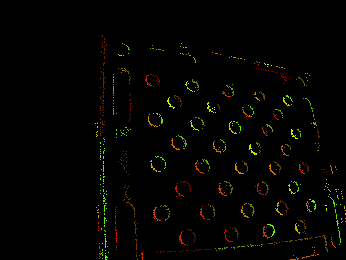}
\includegraphics[width=0.45\linewidth]{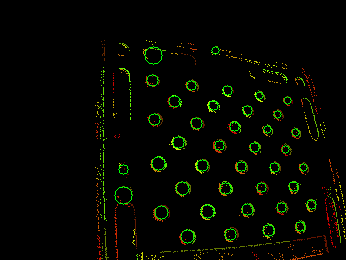}
\includegraphics[width=0.45\linewidth]{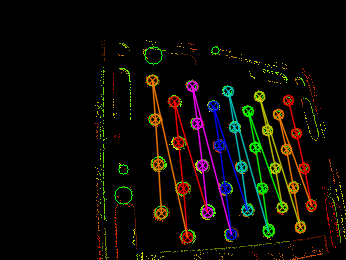}
\caption{The different steps of our pattern detection method: Accumulated event frame (top left), clustering result (top right), circle extraction (bottom left), and pattern detection (bottom right).}
\label{fig:process}
\end{figure}

A high-level overview of our proposed calibration framework is illustrated in Figure \ref{fig:flowchart}. The algorithm groups events into sufficiently small temporal subsets on which we then perform feature extraction and pattern detection. The different steps are illustrated in Figure \ref{fig:process}. $t_{\text{start}}$ and $t_{\text{end}}$ are defined as the timestamps of the first and last event within a temporal subset. Note that the interval duration is gradually increased as long as no pattern is detected and the number of events within the interval remains below a certain threshold. The event accumulation---which we here denote as a \textit{reference frame}---is buffered once the pattern is successfully detected, or cancelled if the total number of events exceeds the said threshold (i.e. no robust detection is expected to happen) or the duration of the interval becomes too long (i.e. not enough motion, pattern not in field of view, etc.). We have
\begin{equation}
    min\cdot\tau \leq |t_{\text{end}}-t_{\text{start}}| \leq max\cdot\tau.
\end{equation}
Note that the minimum gap between two subsequent intervals is furthermore given by $g\cdot\tau$, which causes the frames to be sufficiently distributed. The timestamp given to a reference frame is the center of the interval.

The pattern detection has two steps. The first one is given by the initial circle extraction, the details of which are outlined in Section \ref{sec:feature extraction}. The second step consists of employing the standard calibration pattern detection module from OpenCV \cite{opencv_library}. Once sufficient positive reference frames have been collected, the intrinsic and extrinsic parameters for each reference frame are initialized with an off-the-shelf tool such as the OpenCV implementation of Bouguet's toolbox \cite{bouguet}. Extrinsic parameters for each of the views are notably solved by utilizing a PnP method \cite{collins2014infinitesimal} within a RANSAC scheme with approximate intrinsic parameters obtained by Zhang's method \cite{zhang20}. The initialization is completed by outlier removal and feature cross-validation techniques. The following strategies are applied:
\begin{itemize}
    \item Before adding a reference frame into the buffer, we verify that the orientation of the extracted pattern is sufficiently consistent with previous reference frames. The check is performed by calculating the inscribed angle of the directions of corresponding pattern rows in successive frames divided by the time that has elapsed between those frames (i.e. a measure of the rotational velocity of the pattern).
    \item Once initial extrinsic parameters have been estimated, we approximate translational and angular velocities between successive frames, and again put a threshold on those values.
    \item The feature cross-validation relies on the initial intrinsic and extrinsic parameters to reproject the pattern circles into each reference frame. We start by removing circles that locate outside of the image plane. We then recluster the events based on the newly re-projected circle locations, and refit the circles. To conclude, the newly fitted circles are compared against the projected pattern circles. If the two circles have obvious differences, the corresponding features are again dropped. Reference frames with too few remaining features are dropped as well.
\end{itemize}

Note that after the pattern detection is completed, each surviving event is assigned with exactly one reference frame and exactly one of the circles on the pattern. The last step of the complete calibration procedure is given by an optimization module that directly uses the asynchronous events rather than the fitted circles, the location of which is only approximate owing to the temporal aggregation of the reference frames. The optimization objective consists of a multi-segment continuous time trajectory optimization that jointly optimizes over motion parameters and intrinsic camera parameters by minimizing the geometric distance between back-projected events and their corresponding pattern circles. This optimization is outlined in Section \ref{sec:optimization}.



\subsection{Feature extraction}
\label{sec:feature extraction}

While for conventional cameras, calibration patterns are easily detected from regular images by using standard feature extraction approaches, the nature of event data makes this problem less trivial.  As mentioned in Section \ref{sec:framework_overview}, we start by creating virtual frames called \textit{reference frames}, which represent event accumulations and serve for the feature detection. As long as local optical flow is sufficiently high, the events in their reference frame are given by pairs of clusters of opposing polarity located at opposing poles of a circle. Our feature extraction method aims at identifying the occurrence of such patterns.

\begin{figure}[tb]
\centering
\vspace{0.2cm}
\includegraphics[width=0.45\linewidth]{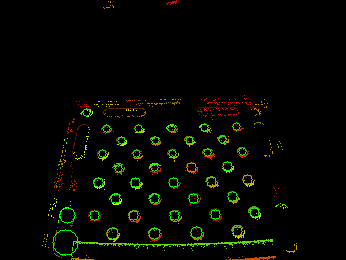}
\includegraphics[width=0.45\linewidth]{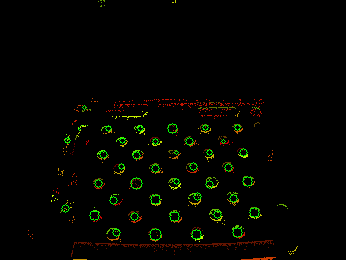}
\caption{\textbf{Left}: \textit{hard feature extraction result}. \textbf{Right}: \textit{soft feature extraction result}.}
\vspace{-0.2cm}
\label{fig:feature_extraction}
\end{figure}

We start by running DBSCAN clustering \cite{ester1996density} individually on the positive and negative polarity subsets of the events. We then remove clusters with too few samples, which are regarded as noise. We also calculate the median of each cluster which is adopted as the corresponding cluster center. The feature extraction mechanism then relies on one of two alternatives:
\begin{itemize}
    \item \textit{Hard feature extraction}: For each cluster of positive events, we search its $\mathit{k}$ nearest neighbors in the negative cluster set, and consider them as potential cluster pairs. Cluster distance is evaluated by simply considering the distance between the cluster centers. We then apply circle fitting \cite{kaasa1976circle} to every potential cluster pair, and afterwards compare the diameter and centre of the fitted circle against the euclidean distance and the midpoint between the two original cluster centers. If passed, we finally evaluate the normalized circle fitting error to select the best cluster pair. This method provides a reliable feature extraction result, but produces less reference frames due to its strict constraints, especially if the camera adopts a more inclined view onto the calibration board, thus causing the features to appear as ellipses rather than circles.
    \item \textit{Soft feature extraction}: In order to deal with noisy scenarios or situations in which the camera adopts a more inclined view onto the calibration pattern, we introduce a simplified feature extraction mechanism. For each positive cluster, we simply find its nearest negative cluster. We then immediately hypothesize circle features by adopting the line between two cluster centers as the diameter. Wrong features are filtered out by checking the normalized circle fitting error. The method provides slightly less stable feature extraction but produces more reference frames in highly challenging scenarios.
\end{itemize}
Note that we also do an inverse search by starting from negative clusters and exploring the nearest neighbours within the positive clusters. A mutual consistency check verifies that features are correctly extracted. The feature detection result for both methods is shown in Figure \ref{fig:feature_extraction}.

\subsection{Multi-segment based optimization}
\label{sec:optimization}

Let there be $N$ events that survived the initial pattern detection stage and let $\mathcal{E} = \{e_{i}\}_{i=1}^{N}$ denote the set of all events\footnote{Note that the events that are considered for the final optimization stage are actually more than the events that have been serving for the initial extraction of the circles. At the beginning of the optimization, we assign further nearby events to circles based on both temporal difference with respect to a reference view as well as geometric distance with respect to a circle center. This procedure approximately doubles the number of considered events while still ensuring computational efficiency of the overall optimization procedure.}. Each event $e_i = \{ \mathbf{m}_i, t_i, b_i \}$ is defined by its pixel location $\mathbf{m}_i$, timestamp $t_i$, and polarity $b_i$. The initial pattern detection stage also returns an ordered set of reference frames $\Phi= \{\mathcal{F}_{j}\}_{j=1}^{M}$ for which initial poses are available. Each of the events is assigned to exactly one reference frame and---more specifically---one of the pattern circles within that frame. Let $f_i$ therefore denote the reference frame of event $e_i$, and $s_i$ the index of its corresponding circle within the pattern. The idea of our optimization framework consists of finding a continuous-time trajectory parametrization that minimizes the distance between the circle corresponding to an event and the intersection point between the pattern plane and the spatial ray corresponding to that event. 

However, it is clear that it is not always possible to robustly detect the calibration pattern, which is why we have to formulate the optimization problem as a function of multiple smooth segments rather than a single long curve. This is especially true as the calibration pattern can only be detected once there is sufficient motion dynamics. To initialize the multiple spline segments, we group the reference views based on temporal proximities. A sub-set is defined by a sequence of subsequent reference views for which the difference between the respective timestamps stays below a certain threshold value. Let there be $L$ segments resulting from this grouping process. We have
\begin{equation}
  \Phi_{l} = \{ \mathcal{F}_{j} \}_{j=a_l}^{b_l}\text{, }l=1,\ldots,L.
\end{equation}
Let $t^{\text{first}}_{a_l}$ and $t^{\text{last}}_{b_l}$ furthermore be the timestamps of the very first and last event in $\Phi_l$, respectively. 

We now proceed to the details of the continuous-time parametrization of the trajectory segments. The various degrees of freedom of each trajectory segment are expressed by a time-parametrized vectorial function, many of which appear in the form of a weighted combination of temporal basis functions. Multiple alternatives exist, such as FFTs, polynomial kernels, or B\'ezier splines. In this work, we use the $p$th-degree B-spline curve
\begin{equation}
	\mathbf{s}_l(u|\mathcal{P}_l) = \sum_{i=0}^{n^l} N_{i,p}(u) \mathbf{p}^{l}_i ,\qquad t^{\text{first}}_{a_l}\leq u\leq t^{\text{last}}_{b_l},
\end{equation}
where $u$ represents the time parameter, $\{\mathbf{p}_i^l\}$ represent the $n^l+1$ control points defining the $l$-th trajectory segment, and $\{N_{i,p}(u)\}$ are the $n^l+1$ $p$th-degree B-spline basis functions defined on the monotonically increasing and non-uniform knot vector $\mathbf{u}^l = \{ \underbrace{t^{\text{first}}_{a_l},\dots,t^{\text{first}}_{a_l}}_{p+1}, u^l_{p+1}, \dots, u^l_{n}, \underbrace{t^{\text{last}}_{b_l},\dots,t^{\text{last}}_{b_l}}_{p+1} \}$. $\mathcal{P}_l=\{\mathbf{p}_i^l\}_{i=0}^{n^l}$ represents the set of the control points of the $l$-th trajectory segment. 

Given the parameter $u$, in order to obtain its exact value on a B-spline curve, we firstly find $u$'s knot span $k$ in the knot vector $\mathbf{u}^l$. $k$ is defined such that $u\in [u^l_k,u^l_{k+1})$, and an exception is given when $u=u^l_{n+1}$ in which case $k=n^l$. We then compute the basis functions $\{N_{i,p}(u)\}$, which are non-zero only if $i=\{k-p, k-p+1,\dots,k\}$. Finally, we multiply the values of the nonzero basis functions with their respective control points, and sum up the terms. Note that the shape of the curve is controlled by the control points only, and the form of a B-spline and the basis functions are generally fixed.

We have not yet defined what the degrees of freedom are. We use the 7-dimensional B-spline curves given by
\begin{equation}
  \mathbf{s}_l(u|\mathcal{P}_l) = \left[ \begin{matrix} \mathbf{t}(u|\mathcal{P}_l) \\ \mathbf{q}(u|\mathcal{P}_l) \end{matrix} \right]\text{, }l=1,\ldots,L,
\end{equation}
where $\mathbf{t}$ is the position of the event camera expressed in a world frame, and $\mathbf{q}$ its orientation as a unit quaternion. For rotation groups, we in fact have two choices to represent it by a B-spline:
\begin{itemize}
    \item The unit quaternion approximation by Kang et al. \cite{kang1999cubic}, which approximates the rotation by a 4-dimensional unit quaternion B-spline.
    \item The SO3 Lie Group B-spline \cite{sommer2016continuous, sommer2020efficient}.
\end{itemize}
Our choice of the first relies on our experiments, which show that there is no notable difference between the quality of results achieved by the two options while the unit quaternion B-spline parametrization is about twice as fast.

In order to use B-splines to parametrize smooth trajectories inside our optimization back-end, we require an appropriate initialization of the control points. We applied an existing spline curve global approximation algorithm \cite{piegl2012nurbs} to initialize each spline segment from a set of samples along the curve. The samples for the $l$-th segment are notably given by the 7-vector representations of the initial absolute poses of the reference frames in $\Phi_l$, which we here denote $\{\mathbf{d}_{a_l},\dots,\mathbf{d}_{b_l}\}$. Let $\{t_{a_l},\ldots,t_{b_l}\}$ furthermore be the time-stamps of these reference frames. We use the automatic knot spacing algorithm described in (9.69) in \cite{piegl2012nurbs} to define an appropriately knot vector $\mathbf{u}^l$. This algorithm guarantees that every knot span contains at least one sample, which causes the linear problem of the initialization to be well-conditioned. Once data points are given, let $\mathbf{p}_0^l = \mathbf{d}_{a_l}$ and $\mathbf{p}_n^l = \mathbf{d}_{a_l}$. The remaining samples are then used to optimize the control point set $\mathcal{P}_l$ in the sense of the least-squares objective
\begin{equation}
    \underset{\mathcal{P}_l}{\arg\min}\sum_{j=a_l+1}^{b_l-1} \|\mathbf{d}_j - \mathbf{s}_l(t_j|\mathcal{P}_l)\|^2.
\end{equation}

We are now ready to formulate our overall optimization objective. Let $\mathcal{S}=\{\mathbf{s}_1(u|\mathcal{P}_1),\ldots,\mathbf{s}_L(u|\mathcal{P}_L)\}$ be the entire set of our continuous-time, 7-dimensional B-spline trajectory segments, and $\Psi=\{\mathcal{P}_1,\ldots,\mathcal{P}_L\}$ be the set of all control point sets. Let $\pi_\mathbf{k}^{-1}(\cdot)$ furthermore be our image-to-camera transformation function, which transforms 2D image points into 3D points on the normalized image plane in the camera frame. $\pi_\mathbf{k}^{-1}(\cdot)$ is a function of the camera's intrinsic parameters $\mathbf{k}$. $\pi_\mathbf{k}^{-1}(\cdot)$ is left unspecified in the below optimization problem and may indeed be exchanged against any continuously differentiable normalization function of $\mathbf{k}$. We use a perspective camera with radial distortion as an example, in  which case the parameter vector would be given by $\mathbf{k} = \left[f_x, f_y, c_x, c_y, k_1, k_2, k_3, k_4, k_5 \right]^T$. It includes parameters of an inverse lens distortion function. We use the higher-order polynomial based inverse radial distortion model of Drap et al. \cite{drap2016exact}, which contradicts the common usage of a distortion function. Note however that both models can always be converted into one-another. For an event $e_i$, the normalization is then given by
\begin{equation}
    \pi_\mathbf{k}^{-1}(\mathbf{m}_i) = \begin{bmatrix}\beta P_x & \beta P_y & 1\end{bmatrix}^T
\end{equation}
where 
\begin{equation}
   \begin{split}
    \mathbf{m}_i &= \begin{bmatrix} p_x & p_y \end{bmatrix}^T \,, \\
    P_x &= \left(p_x - c_x\right)/f_x \,, \\ 
    P_y &= \left(p_y - c_y\right)/f_y \,, \\
    \alpha &= \sqrt{P_x^2 + P_y^2} \,, \\ 
    \beta &= 1 + k_1\alpha^2 + k_2\alpha^4 + k_3\alpha^6 + k_4\alpha^8 + k_5\alpha^{10} \,.
   \end{split} 
\end{equation}

Knowing that the world frame is defined such that the calibration coincides with the plane $z=0$, we may now calculate the intersection point between the pattern plane and the point along the spatial ray defined by the normalized event location and the pose of the camera at the specific time the event was fired. We define this point by $\lambda_i$, the depth of $\pi_\mathbf{k}^{-1}(\mathbf{m}_i)$. It is given by
\begin{equation}
\lambda_i = \frac{-(\mathbf{t}(t_i|\mathcal{P}_{f_i}))_{z}}{\text{row}(\mathbf{R}(\mathbf{q}(t_i|\mathcal{P}_{f_i})), 3) \pi_\mathbf{k}^{-1}(\mathbf{m}_i)},
\end{equation}
where $\mathbf{T}(t_i|\mathcal{P}_{f_i}) =\left[\begin{matrix}\mathbf{R}(\mathbf{q}(t_i|\mathcal{P}_{f_i})) & \mathbf{t}(t_i|\mathcal{P}_{f_i}) \\ \mathbf{0}^\intercal & 1 \end{matrix}\right]$ transforms the normalized point from the camera frame to the world coordinate frame defined by the 3D calibration pattern. $(\mathbf{t}(t_i|\mathcal{P}_{f_i}))_{z}$ is the third element of the position $\mathbf{t}$, and $\text{row}(\cdot,3)$ takes the third row of the rotation matrix. Note that $f_i$ is redefined as the index of the trajectory segment to which event $e_i$ belongs.

The calibration optimization objective jointly optimizes over the intrinsic parameters as well as the control points of each trajectory segment. It is finally given by:
\begin{gather}\label{eq:finalEq}
\min_{\mathbf{k}, \Psi} \sum_{i} \rho\left(\big\| \|\mathbf{x}_i - \mathbf{l}_{s_i}\| - r \big\|^{2}\right) \\
\text{where } \mathbf{x}_i = \lambda_i\mathbf{R}(\mathbf{q}(t_i|\mathcal{P}_{f_i}))\pi_\mathbf{k}^{-1}(\mathbf{m}_i)+ \mathbf{t}(t_i|\mathcal{P}_{f_i}). \nonumber
\end{gather}
$\mathbf{x}_i$ represents the 3D location of event $e_i$ in the pattern plane, $r$ the pattern circle radius, and $\mathbf{l}_{s_i}$ the 3D location of the corresponding pattern circle center (recall that $s_i$ defines the index of the circle that event $e_i$ corresponds to). Note that the third coordinate of $\mathbf{l}_{s_i}$ is always zero. $\rho\left(\cdot\right)$ is a loss function to mitigate the influence of outliers (e.g. Huber loss).



\section{Experimental evaluation}
\label{sec:experiments}
We introduce further details about the implementation of our approach and test our methods on multiple real datasets. We assess both quality of the estimated intrinsic parameters over different types of lenses, as well as the accuracy of the estimated extrinsic parameters compared against groundtruth trajectories.

\subsection{Experiment setup}
We evaluate the performance of our method by using a DAVIS346 event camera with different types of lenses as listed in Table \ref{tab:lenses}. The camera has a resolution of 346$\times$260. The produced event stream has a maximum temporal resolution of 1$\mu$s. The camera has the advantage of also capturing regular frames at a frame rate of 30Hz under regular illumination conditions, which lets us easily compare our method against a high-quality, regular image-based calibration method\footnote{We do not compare against the calibration methods from \cite{ev1,ev2,ev3} because they effectively emulate the traditional calibration process by keeping the calibration pattern still and by generating events through the flashing pattern. Hence their achievable calibration accuracy can be considered equivalent to the one obtained by traditional calibration methods for regular cameras.} such as the open-source OpenCV calibration pipeline. The calibration pattern used throughout the evaluation is a $9\times4$ asymmetric circle pattern. All our experiments are conducted on a desktop with 32GB RAM and an Intel Xeon 3.2 GHz CPU. Implementations are made in C++, and use OpenCV~\cite{opencv_library}, Eigen~\cite{eigen}, and the Ceres optimization toolbox~\cite{ceres-solver} with automatic differentiation. In order to quantitatively evaluate the performance of our calibration result, we evaluate the absolute trajectory error~(ATE) between the estimated extrinsic parameters and its groundtruth trajectories by utilizing the tools from the TUM-RGBD\cite{sturm12iros} benchmark. Groundtruth is provided by an Optitrack external motion tracking system.
\begin{table}[htb]
\centering
\vspace{0.2cm}
\caption{Specifications of the lenses used in our experiments. FOV represents the field-of-view of the lens.}
\resizebox{\linewidth}{!}{%
\begin{tabular}{|c|c|c|c|}
\hline
Name   & Label           & FOV  & TV Distortion \\ \hline
\textbf{Lens-1} & HIK-MVL-MF1220M & 40.2 & -1.01         \\ \hline
\textbf{Lens-2} & Kowa-LM5JCM     & 82.4 & -0.50         \\ \hline
\textbf{Lens-3} & Kowa-LM6JC      & 81.9 & 10.70         \\ \hline
\end{tabular}%
}
\vspace{-0.2cm}
\label{tab:lenses}
\end{table}

\begin{figure}[htb]
\centering
\includegraphics[width=0.45\linewidth]{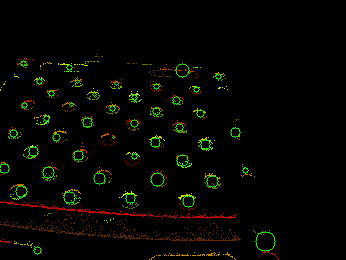}
\includegraphics[width=0.45\linewidth]{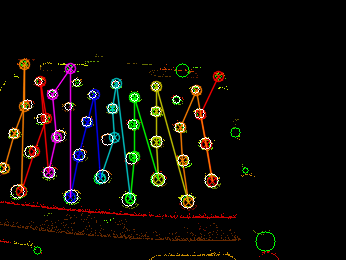}\\
\includegraphics[width=0.45\linewidth]{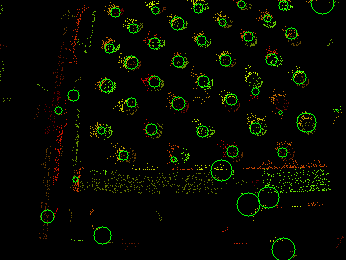}
\includegraphics[width=0.45\linewidth]{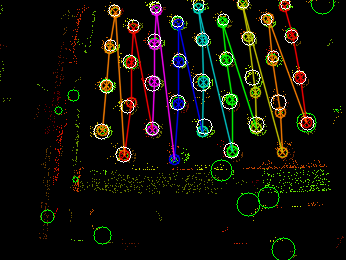}\\
\includegraphics[width=0.45\linewidth]{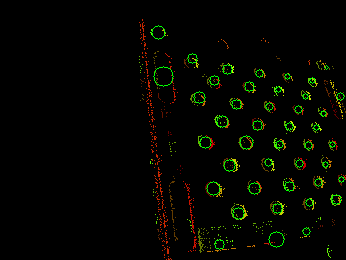}
\includegraphics[width=0.45\linewidth]{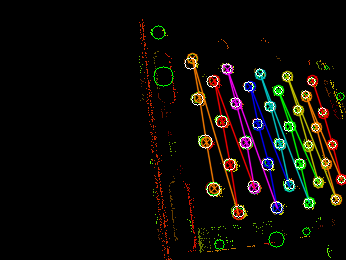}
\caption{Effect of our feature rectification strategy (features are colored in white). \textbf{Left}: Original feature extraction. \textbf{Right}: Rectified features for poor-quality reference frames.}
\label{fig:pattern_refinement}
\end{figure}

\begin{table*}[htb]
\centering
\vspace{0.2cm}
\caption{Comparison against GT. OpenCV's distortion parameters are $[k_1,k_2,p_1,p_2,k_3]$, and the relation to the inverse radial distortion model is explained in \cite{drap2016exact} ($k_1$ has opposite sign).}
\resizebox{\textwidth}{!}{%
\begin{tabular}{|c|c|c|c|c|c|c|c|c|c|c|c|}
\hline
\multirow{2}{*}{Dataset} & \multirow{2}{*}{Duration(s)} & \multirow{2}{*}{FrameNum} & \multicolumn{4}{c|}{ATE (cm)} & \multirow{2}{*}{$f_x$} & \multirow{2}{*}{$f_y$} & \multirow{2}{*}{$c_x$} & \multirow{2}{*}{$c_y$} & \multirow{2}{*}{$k_1$} \\ \cline{4-7}
 &  &  & rmse & mean & median & std &  &  &  &  &  \\ \hline
lens1-slow1-image & \multirow{3}{*}{110} & 260 & \textbf{0.5470} & \textbf{0.4660} & \textbf{0.4146} & \textbf{0.2864} & 345.08 & 345.24 & 167.79 & 123.88 & -0.3695 \\
lens1-slow1-event-soft &  & 2136 & 0.9305 & 0.7283 & 0.6047 & 0.5792 & 347.99 & 348.90 & 166.43 & 121.44 & 0.3694 \\
lens1-slow1-event-hard &  & 1412 & 1.0338 & 0.8071 & 0.6659 & 0.6460 & 349.31 & 350.06 & 165.77 & 118.97 & 0.3837 \\ \hline
lens1-slow2-image & \multirow{3}{*}{97} & 207 & \textbf{1.1587} & \textbf{0.9053} & \textbf{0.8582} & \textbf{0.7231} & 342.53 & 342.78 & 168.056 & 126.065 & -0.3675 \\
lens1-slow2-event-soft &  & 1448 & 1.9136 & 1.1268 & 0.9491 & 1.5467 & 344.95 & 344.84 & 167.12 & 121.83 & 0.3662 \\
lens1-slow2-event-hard &  & 1859 & 1.3622 & 1.0241 & 0.8809 & 0.8982 & 344.85 & 344.755 & 166.97 & 121.95 & 0.3598 \\ \hline
lens2-slow-image & \multirow{3}{*}{94} & 131 & 11.278 & 2.7100 & 1.2878 & 10.948 & 272.18 & 272.75 & 162.558 & 124.134 & -0.0706 \\
lens2-slow-event-soft &  & 993 & 1.4796 & 1.3798 & 1.3607 & 0.5343 & 269.674 & 272.135 & 163.465 & 128.186 & 0.069 \\
lens2-slow-event-hard &  & 438 & \textbf{1.2254} & \textbf{1.1257} & \textbf{1.0515} & \textbf{0.4840} & 270.84 & 273.44 & 163.04 & 129.51 & 0.0440 \\ \hline
lens3-slow-image & \multirow{3}{*}{69} & 127 & 1.9512 & 1.5330 & 1.3765 & 1.2071 & 338.586 & 338.013 & 159.87 & 116.66 & -0.3258 \\
lens3-slow-event-soft &  & 902 & \textbf{1.1889} & 0.9964 & \textbf{0.9082} & \textbf{0.6487} & 334.377 & 335.8 & 162.076 & 118.978 & 0.34097 \\
lens3-slow-event-hard &  & 480 & 1.3162 & \textbf{0.9768} & 0.9177 & 0.8822 & 335.356 & 336.636 & 161.1 & 118.96 & 0.3499 \\ \hline
lens1-fast1-image & \multirow{3}{*}{8.16} & 37 & 0.7605 & 0.6851 & 0.6162 & \textbf{0.3302} & 342.82 & 342.47 & 170.39 & 124.999 & -0.3661 \\
lens1-fast1-event-soft &  & 101 & 0.8362 & 0.7430 & 0.6808 & 0.3835 & 345.097 & 348.76 & 171.94 & 122.31 & 0.4093 \\
lens1-fast1-event-hard &  & 76 & \textbf{0.7215} & \textbf{0.5887} & \textbf{0.5002} & 0.4172 & 346.18 & 350.725 & 172.72 & 122.56 & -0.0146 \\ \hline
lens1-fast2-image & \multirow{3}{*}{8.4} & 84 & 0.8774 & 0.7277 & 0.6033 & 0.4902 & 337.234 & 338.36 & 169.83 & 128.125 & -0.3637 \\
lens1-fast2-event-soft &  & 331 & 0.8722 & 0.6505 & 0.4813 & 0.5810 & 345.93 & 348.73 & 164.76 & 126.233 & 0.34578 \\
lens1-fast2-event-hard &  & 244 & \textbf{0.6025} & \textbf{0.4695} & \textbf{0.3751} & \textbf{0.3776} & 340.765 & 343.03 & 168.158 & 125.877 & 0.2943 \\ \hline
lens1-blur1-image & \multirow{3}{*}{70} & 4 & 3.170 & 2.7338 & 2.3104 & 1.6047 & 341.466 & 334.28 & 188.9 & 122.555 & -1.3176 \\
lens1-blur1-event-soft &  & 1068 & 1.12 & 0.7524 & 0.6442 & 0.8296 & 344.99 & 347.486 & 158.6 & 120.40 & 0.3856 \\
lens1-blur1-event-hard &  & 176 & \textbf{0.7456} & \textbf{0.639} & \textbf{0.5718} & \textbf{0.3842} & 343.115 & 345.94 & 160.09 & 121.3 & 0.3262 \\ \hline
lens1-blur2-event-soft & 6.537 & 70 & \textbf{1.409} & \textbf{1.13} & \textbf{0.931} & \textbf{0.8417} & 359.098 & 365.023 & 165.022 & 123.3 & 0.5227 \\ \hline
\end{tabular}%
}
\label{tab:gt-compare}
\end{table*}

\begin{figure*}[htb]
    \centering
    \includegraphics[width=0.225\linewidth]{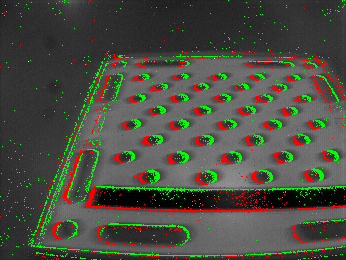}
    \includegraphics[width=0.225\linewidth]{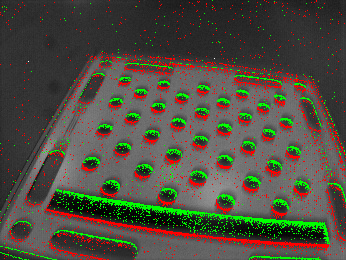}
    \includegraphics[width=0.225\linewidth]{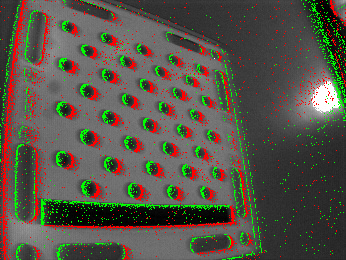}
    \includegraphics[width=0.225\linewidth]{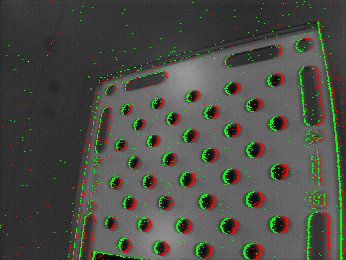}
    \includegraphics[width=0.225\linewidth]{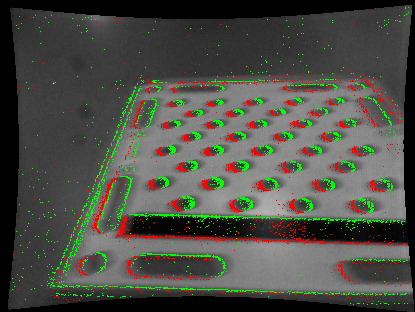}
    \includegraphics[width=0.225\linewidth]{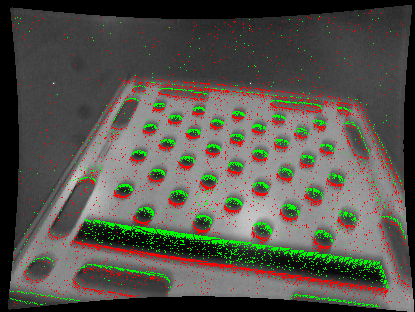}
    \includegraphics[width=0.225\linewidth]{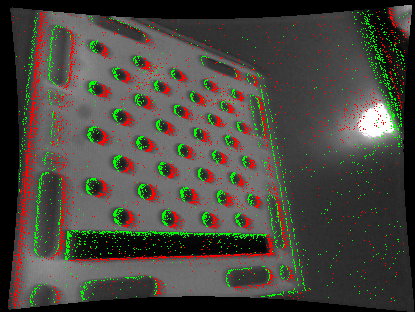}
    \includegraphics[width=0.225\linewidth]{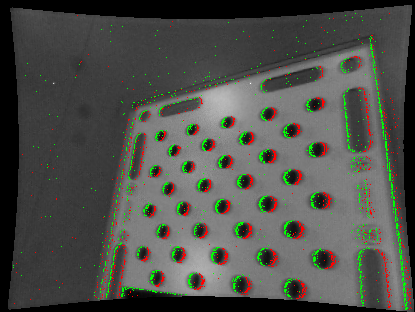}
    \caption{Undistortion of events and images using our calibration result on \textit{1104-fisheye1}.}
    \label{fig:undistortion}
    \vspace{-0.4cm}
\end{figure*}

\subsection{Accuracy evaluation}
We test the methods on several real sequences with different lenses, motion characteristics, and illumination conditions. For different lens types, the sequences are named \textit{lensx-slowx}. They are captured under normal conditions without highly dynamic motion or challenging illumination. Next, we perform high speed calibration tests with sequences shorter than 10s, which are called \textit{lens1-fast1} and \textit{lens1-fast2}. They are captured with \textit{lens1} still under normal illumination conditions. We conclude with long and short sequences under high speed motion as well as reduced illumination, which leads to noisy event sequences and blurry images. The latter sequences are named \textit{lens1-blur1} and \textit{lens1-blur2}.

Note that intrinsic camera parameters are potentially inter-correlated, which is why it is not possible to regard the regular camera alternative as ground truth. In order to fairly assess the performance of our methods, we perform a quantitative comparison of the absolute trajectory error between our proposed event-based solution and the regular camera alternative. Table \ref{tab:gt-compare} summarizes all results. We evaluate the root-mean-square~(\textbf{rmse}), mean~(\textbf{mean}), median~ (\textbf{median}) and standard deviation~(\textbf{std}) errors between the estimated extrinsics and ground truth. As can be easily observed, our method produces similar calibration results to OpenCV while showing higher pose accuracy than the standard image-based calibration tool. The following is worth noting:
\begin{itemize}
    \item \textbf{Calibration for different lenses:} We apply our calibration to different lenses in sequences \textit{lens1-slow1}, \textit{lens1-slow2}, \textit{lens2-slow}, and \textit{lens3-slow}. The obtained results are comparable to the normal image alternative. Note that---while our method always returns low ATE errors in absolute terms---our motion parameterization works best in more dynamic scenarios. This explains the small gap to regular camera alternatives on \textit{lens1-slow1} and \textit{lens1-slow2}.
    
    \item \textbf{Comparison of different circle extraction schemes:} Table \ref{tab:gt-compare} shows that the \textit{hard feature extraction} method is better than \textit{soft feature extraction} in terms of the ATE error. However, it has difficulties to correctly identify distortion coefficients when there are only few reference frames (cf. $k_1$ values on \textit{lens1-fast1}). \textit{Hard feature extraction} has a preference for noise-less reference frames which only distribute a part of the whole trajectory, which may cause a bad distribution of the reference views especially if there are only few of them.
    
    \item \textbf{High-speed calibration:} \textit{lens1-fast1}, \textit{lens1-fast2}, and \textit{lens1-blur2} are high-speed sequences of less than 10s duration. Our method works well even on such short sequences.
    
    \item \textbf{High-speed calibration under bad illumination conditions:} \textit{lens1-blur1} and \textit{lens1-blur2} are high-speed datasets captured under unfavourable illumination conditions. It produces blurry images and noisy events, and renders calibration from normal images unusable. In contrast, our proposed methods still works well, especially if using the \textit{soft feature extraction} strategy.

\end{itemize}

\subsection{Further results}

The effectiveness of our feature rectification approach is visualized in Figure \ref{fig:pattern_refinement}. White circles indicate rectified features whereas colored ones represent the originally detected pattern. Outlier features are rectified and moved to correct positions, while badly observed features are removed. As demonstrated in the figure, our feature rectification method handles various scenarios such as mismatched features~(\textbf{top}), noisy event streams~(\textbf{center}), and heavily inclined views onto the pattern causing the features to appear in elliptical shape~(\textbf{bottom}). Figure \ref{fig:undistortion} finally visualizes undistorted images from different perspectives. As can be observed, our method produces visually pleasing results.


\section{Conclusion}

We present a novel framework for event camera calibration that requires only traditional calibration patterns. The pattern is rendered observable as soon as the camera is under motion. Multiple trajectory segments are taken jointly into account. They are optimized in continuous time which permits every event to be accounted for with its exact time stamp. By adding several techniques to ensure the quality of the correspondences between events and pattern features, the overall framework finally achieves satisfying results that are comparable to regular camera alternatives. We will release our framework and are confident that it will benefit future research on event-based 3D vision.

{\small
\bibliographystyle{IEEEtran}
\bibliography{root}
}
\end{document}